\documentclass[sigconf]{acmart}
\AtBeginDocument{%
  \providecommand\BibTeX{{%
    \normalfont B\kern-0.5em{\scshape i\kern-0.25em b}\kern-0.8em\TeX}}}

\setcopyright{acmcopyright}
\copyrightyear{2023}
\acmYear{2023}
\acmDOI{XXXXXXX.XXXXXXX}

\acmConference[CONCATENATE - HRI]{}{March 13, 2023}{Stockholm, SE}
\acmBooktitle{HRI '23: ACM/IEEE International Conference on Human-Robot Interaction, March 13--16, 2023, Stockholm, SE} 
\acmPrice{}
\acmISBN{}

\usepackage{soul}
\setstcolor{red}

\newcommand{\reconsider}[1]{\textcolor{black}{#1}}
\newcommand{\rephrase}[1]{\textcolor{black}{#1}}
\newcommand{\revisit}[1]{\textcolor{black}{#1}}

\newcommand{\revised}[1]{\textcolor{black}{#1}}
\begin{document}

\title{Modeling Group Dynamics for Personalized Robot-Mediated Interactions}
%\title{Human Understanding Models for Group Interactions}

\author{Hifza Javed}
\orcid{0000-0002-5414-6318}
\affiliation{%
  \institution{Honda Research Institute USA, Inc.}
  \streetaddress{70 Rio Robles}
  \city{San Jose}
  \state{California}
  \country{USA}
  \postcode{95134}
}
\email{hifza_javed@honda-ri.com}

\author{Nawid Jamali}
\affiliation{%
  \institution{Honda Research Institute USA, Inc.}
  \streetaddress{70 Rio Robles}
  \city{San Jose}
  \state{California}
  \country{USA}
  \postcode{95134}
}
\email{njamali@honda-ri.com}

\renewcommand{\shortauthors}{Javed and Jamali.}

\begin{abstract}
    The field of human-human-robot interaction (HHRI) uses social robots to positively influence  how humans interact with each other. This objective requires models of human understanding that consider multiple humans in an interaction as a collective entity and represent the group dynamics that exist within it. Understanding group dynamics is important because these can influence the behaviors, attitudes, and opinions of each individual \revised{within the group, as well as} the group as a whole. Such an understanding is also useful when personalizing an interaction between a robot and the humans in its environment, where a group-level model can facilitate the design of robot behaviors that are tailored to a given group, the dynamics that exist within it, and the specific needs and preferences of the individual interactants.
    In this paper, we highlight the need for group-level models of human understanding in human-human-robot interaction research \revised{and how these can be useful in developing personalization techniques}. We survey existing models of group dynamics and categorize them into models of social dominance, affect, social cohesion, and conflict resolution. We highlight the important features these models utilize, evaluate their potential to capture interpersonal aspects of a social interaction, \revised{and highlight their value for personalization techniques}. Finally, we identify \revised{directions for future work}, and make a case for models of relational affect as an approach that can better capture group-level understanding of human-human interactions and be useful in personalizing human-human-robot interactions.
\end{abstract}



%\received{}
%\received[revised]{}
%\received[accepted]{}

\maketitle

\section{Introduction}
\rephrase{As robots become more integrated into our everyday lives, they will be sharing space with humans. The way these robots behave will affect social and interpersonal dynamics among the humans in their environments. Existing social robotics literature often involves the interaction of an individual human with a robotic agent, and explores the use of social robots in different roles that may be executed with various levels of personalization. However, modeling interactions between a group of humans presents complex challenges that have not been studied to the same extent and are unique to multiparty interactions---such as the study of social dynamics, sub-group formation, evaluation of group affect, and mediation---that need to take multiple interactants’ perspectives under consideration. This has led to the emergence of a sub-field called human-human-robot interaction (HHRI)} that focuses on  how humans interact with each other in the presence of a robot~\cite{oliveira2021human, wisowaty2019group}.

One of the main goals of HHRI research is social mediation~\cite{birmingham2020can, papadopoulos2012exploring, druckman2021best, tuncer2022robot}, in which a robot produces behaviors that positively influence how humans in its environment interact with each other~\cite{oliveira2021towards}. Human-human interactions involve multiple individuals that interact with one another through complex verbal and nonverbal signals that are contextual and change over time~\cite{argyle1972non}. Successful mediation of such interactions would require that the robot be able to recognize human signals in real-time and understand what these mean in a given context. This understanding can then be used to generate robot behaviors that may help build relationships, improve connectedness, and achieve the group's specific goals. \revisit{Personalizing robot-mediated interactions that involve multiple humans becomes even more challenging given that the robot must acquire an understanding of the individual interactants in the group, as well as an understanding of the group as a whole, so as to allow its behaviors to adapt to the needs and preferences of the interactants in the context of existing social dynamics.}

Personalization has been shown to be an important factor in creating effective robot-mediated interactions for a variety of applications~\cite{lee2012personalization, szafir2012pay, ahmad2017systematic, leyzberg2014personalizing, dautenhahn2004robots}. However, before a robot can provide personalized mediation for group interactions, it must first be equipped with the capability to interpret and evaluate group dynamics.
Group dynamics are defined as the influential actions, processes, and changes that occur within and between group interactants \cite{forsyth2018group}. 
\revised{From the field of social psychology, we know that group-level processes are fundamentally different from individual-level processes~\cite{abrams2020c} and that a group can influence an individual’s behavior due to the individual’s interaction with other group members in a social setting~\cite{lewin1951field}.} Therefore, understanding group dynamics is important because these can influence the behaviors, attitudes, and opinions of each individual within the group and the group as a whole \cite{lee2014group}. 
\revised{Computational methods for modeling human understanding often utilize work from the field of affective computing, where arousal-valence models of emotion \cite{russell1980circumplex} are frequently used to represent Ekman's basic emotions \cite{ekman1992argument}. Often, their aim may be to accurately recognize the observable expressions of these emotions, \cite{jung2017affective}, typically manifested in facial expressions.} This can lead to an emphasis on human understanding models that yield individual-level arousal and valence estimates and focus on the intrapersonal, experiential aspects of affect. Such models
\revised{may not be able to fully represent }the interpersonal or relational aspects of affect that are intrinsic to social interactions between humans \cite{jung2017affective}. As a result, models that represent multiple humans as a collective are not 
\revised{explored to the same extent as individual-level models}. \revisit{This also has implications for personalization frameworks that consume the output of such human understanding models to generate suitable robot behaviors. If these models do not adequately represent the relational aspects of a human-human interaction, personalization techniques will be unable to leverage this critical understanding of group dynamics to adapt its mediation behaviors to the needs of the individual interactants within the context of existing group dynamics.} \revised{In general, personalization methods may benefit from a group-level view of human-human interactions where a combination of individual-level and group-level evaluations may be used to provide mediation that is specifically adapted to the needs and preferences of the interactants within the context of existing group dynamics. This may lead to a highly-tailored and potentially more beneficial  experience for the interactants within the group.}

In this paper, we survey current models of group dynamics utilized in HHRI studies and analyze their ability to represent the interpersonal aspects of a social interaction. \revised{We also highlight how such models may contribute to improvements in personalized interactions with a robot.} \revised{Additionally, we identify directions for future research and make a case for models of relational affect, which can offer metrics that can not only be used directly to assess group states but can also be useful in designing highly personalized human-human-robot interactions.}

\section{Existing Perspectives on Group-Level Modeling}
Previous studies have taken several distinct perspectives on group dynamics. A number of approaches have been explored that target goals like balancing participation from all interactants, improving turn-taking, increasing likeability between group members, and improving overall team performance. We use these goals to categorize the approaches into models of social dominance, affect, social cohesion, and conflict resolution. We survey these models to identify important behavioral cues and other features that can be used to represent various aspects of an interaction between humans, and determine their potential to capture the interpersonal aspects of a social interaction. 
\revised{Additionally, we also highlight how group-level models of interpersonal dynamics can provide valuable insights to contribute to a highly personalizable robot-mediated interaction experience.}

\subsection{Social Dominance}
Social dominance is defined as a relational, behavioral, and interacting state that reflects the actual achievement of influence or control over another via communicative actions \cite{strohkorb2015classification}. Among children, it is found that highly dominant children gain benefit from social attention that other children do not, which can potentially impact their performance in group learning environments \cite{strohkorb2015classification}. At the workplace, it has been found to be associated with harsh power tactics \cite{aiello2018social}. Social dominance is communicative in nature, making it easier to detect, measure, and classify compared to other phenomena. As a result, it has been investigated more thoroughly than the remaining three categories discussed in this paper. 

Visual activity is found to be a strong indicator of dominance. For example, dominant individuals are found to move more often and with a wider range of motion compared to non-dominant ones~\cite{jayagopi2009modeling}. Gesturing while speaking is also found to be positively correlated with high dominance~\cite{jayagopi2009modeling}. It is important to note that patterns of dominance are different for adults and children. Verbal behaviors are better indicators of social dominance in adults whereas non-verbal features are more suitable for children\cite{strohkorb2015classification}. Coercive or aggressive behaviors are more common in younger dominant children (over 5 years old) and prosocial behaviors are more common in elder dominant children \cite{strohkorb2015classification}. Additionally, individuals with high social dominance spend less time looking at others while listening to them compared to individuals lower in social dominance \cite{strohkorb2015classification,dunbar2005perceptions}. From mixed-gender studies, it is found that females speak more than males but their total speaking time and overlap in speech decrease when paired with a male than another female interactant~\cite{skantze2017predicting}.

Important audiovisual features used to model dominance include the length of speaking time, speech loudness, tempo, pitch, and vocal control~\cite{strohkorb2015classification, jayagopi2009modeling, smith2015real}, as well as body movement, posture, elevation, gestures, facial expressions, gaze, and looking-time measures (i.e., the amount of time spent looking at other interactants) \cite{strohkorb2015classification, jayagopi2009modeling}. The presence of verbal and non-verbal backchanneling behaviors ranging from a simple glance to a verbal "mm-hmm"~\cite{tennent2019micbot} are found to be directly and indirectly related to group participation~\cite{tennent2019micbot}. \revised{These approaches that model social dominance often focus on the identification of the least or most dominant interactant within the group, such as the machine-learning based classification of the most dominant individual in a group meeting~\cite{jayagopi2009modeling}, the least and most dominant child in a child-robot interaction~\cite{strohkorb2015classification}, or the least actively participating student in a group project discussion~\cite{tennent2019micbot}. This creates an emphasis on an individual-level view of the interactants only on the extreme ends of the dominance spectrum, rather than a group-level dominance model of the interaction as a whole that captures the balance of participation prevailing within a group interaction.}

\revised{From the perspective of personalizing interactions with a robot, such a group-level view of social dominance can provide uniquely valuable insights. For example, lack of participation from a student in a group discussion where the group dynamics are predominantly healthy may require that a robot occasionally prompt the student to share their views~\cite{gillet2021robot}, as in the case of the rotating Micbot \cite{tennent2019micbot} that targets a balance in participation by simply changing its direction to encourage turn-taking from different interactants. On the contrary, if the group includes some domineering students who frequently dominate the discussion, the interpersonal relationships may suffer. In such a case, if the robot were to urge the quiet, shy student to participate, it may potentially place undue pressure on them and cause them discomfort. Instead, an appropriate approach to balance participation in this case may involve devising situations that encourage the domineering students to self-reflect. For example, the robot may take a domineering comment from a student and rephrase it in a positive way such that it adds value to the discussion without discouraging other students. A robot modeling appropriate social behavior in this manner may encourage the students to follow suit. In this way, an understanding of social dominance within the group may possibly be used to inform the subsequent behavior generation process for the robot in order to enable a higher level of personalization in the interaction.}

\subsection{Affect}
This category encompasses models based on affect, emotions, and/or mood. Affect is an umbrella term in psychology that refers to the experience of feelings, emotions, or moods~\cite{frijda1986emotions}. Emotion is a short-term, intense affective state, associated with specific expressive behaviors; mood is a long-term, diffuse affective state, without such specific behaviors. Mood emphasizes a stable affective context, while emotion emphasizes affective responses to events \cite{xu2015mood}. Naturally, evaluations of group affect are more challenging than dominance since affect indicates an internal feeling state, the expressions of which may not always be unambiguous. \revised{Barsade~\cite{barsade2002ripple} showed the importance of affective behaviors in group interactions by demonstrating that the emotional dynamics and performance of a team can be influenced by the mood of a single member of the team.}

Research on affect recognition frequently employs the valence-arousal model of emotion and utilizes facial expressions-based classification of emotions using deep learning methods~\cite{sharma2019automatic}. For group-level assessment of affect using video-based methods, two types of approaches can be found in literature: bottom-up and top-down. \revised{A bottom-up approach determines the individual affective state of each interactant and combines these to form a group-level measure of affect. How the individual states are combined to yield a group measure is an important factor. To this end, an approach to determine group mood from images explored a number of techniques to reach a group-level affect estimate, including the use of a mean of the face-level estimations, the use of a distribution of all facial expressions as a feature vector, and assigning a variable level of significance to faces based on the size and distance from the group~\cite{vonikakis2016group}.} In contrast, a top-down approach focuses on extracting global, scene-level features to estimate group affect~\cite{sharma2019automatic}.

\revised{Some previous research utilizes a combination of the two approaches to evaluate group affect from image data. Mou et. al.~\cite{mou2015group} used a combination of facial, body, and context features like relative location and size of bodies and faces to train a classifier to classify group affect. Tan et. al.~\cite{tan2017group} used whole images and individual face images as inputs for separate deep convolutional neural networks and then combined the results of both approaches to classify group affective state. Guo et. al.~\cite{guo2017group} used a similar approach for scene-level information but utilized separate classifiers for face and skeleton classifications, fusing the two types of classifications to reach a final estimate.}

Studies that classify emotion dynamics in conversations also utilize features related to utterances and context~\cite{yang2021emotion}, such as audio features that include energy, logarithmic harmonic-to-noise ratio, spectral harmonicity, and psychoacoustic spectral sharpness~\cite{sharma2019automatic}.

An understanding of the group affective state can facilitate a personalized mediation strategy. If a robot is mediating a conflict between two individuals, both of whom are emotionally aroused, a suitable mediation strategy may be to acknowledge their emotions, show empathy, and encourage perspective-taking from both individuals to help regulate emotions~\cite{rupp2008customer}.

\subsection{Social Cohesion}
Social cohesion is defined as "a dynamic process which is reflected in the tendency for a group to stick together and remain united in the pursuit of its instrumental objectives and/or for the satisfaction of member affective needs"~\cite{carron1985development}. It may manifest in the form of interpersonal attraction between group members, group pride, commitment to the task of the group, etc.~\cite{forsyth2018group}. \revised{According to Carron et al.~\cite{carron2000cohesion}, cohesion is a dual construct that exists both on an individual level and on a group level. At the individual level, cohesion refers to a member's perception and evaluation, such as their attraction towards the group, while at the group level, cohesion encompasses factors like team members' perceptions of similarity within the group and their closeness to the entire group~\cite{abrams2020c}. Additionally, for each level, two further dimensions of cohesiveness are defined: task-level cohesiveness and interpersonal cohesiveness~\cite{abrams2020c}. The positive consequences of social cohesion are more participation, high productivity, more success, and more personal level satisfaction~\cite{martens1971group}.}

\revised{Both in dyadic and multiparty human interactions, audiovisual cues have been widely utilized to measure cohesion. Audio cues have been found to be useful indicators of interpersonal synchrony~\cite{hung2010estimating} derived from turn-taking behaviors (pauses between individual turns, turn exchanges, turn length, overlapping turns, etc.), as well as features of vocal prosody (tone and pitch). Campbell et. al~\cite{campbell2008individual} utilized features of speaking activity within dyadic interactions to estimate the degree of synchrony and rapport between the two interactants. A correlation between body motion and speaking activity was also found within four-person conversations~\cite{campbell2008multimodal}, indicating that visual activity can also be informative. For multiparty interactions, visual information such as coarse head and body motion, and body pose information were used along with audio features to determine group interest levels in meetings~\cite{gatica2005detecting}.}

In addition, measures of social cohesion are often obtained from post-interaction surveys, such as the group environment questionnaire~\cite{brawley1985development} to extract members' perception of group cohesion, as well as other surveys to capture interpersonal relationships and likeability \cite{jung2020robot, strohkorb2016improving}.

\revised{Understanding group cohesion may be valuable for personalizing interactions with a robot. Since task-level cohesion is correlated with team performance~\cite{gachter2022measuring}, such an understanding could be utilized by a robot to motivate teams for higher performance in a group task. As an example, consider the application of a social robot to conduct physical rehabilitation sessions for a group of older adults~\cite{shao2023long}. Such robots typically provide positive feedback and praise to encourage intrinsic motivation in the participants~\cite{fasola2012using}. In such a case, if group cohesion is found to be low, the robot may be able to incorporate strategies to improve group cohesion alongside its usual motivational techniques to offer a more tailored rehabilitation program to the group of adults.}

\subsection{Conflict Resolution}
Conflict resolution is the process that two or more parties use to find a peaceful solution to their dispute. It can be important at the workplace where successfully resolving conflicts can lead to greater efficiency and goal achievement, and maintain a positive, comfortable environment for all employees. \revised{Two types of conflicts can occur within teams: relationship and task conflicts~\cite{jehn1995multimethod}. Relationship conflicts pertain to interpersonal incompatibilities involving tension, animosity, and annoyance among the group members, whereas task conflict is related to the substance of the task at hand, such as differences in perspectives and ideas~\cite{jehn1995multimethod}. Both types of conflict can negatively impact team performance~\cite{de2003task}.}

Conflict resolution is widely studied within the context of child-play scenarios~\cite{shen2018stop}, where play behaviors may be categorized into nonsocial constructive play, social constructive play, nonsocial nonconstructive play, and social nonconstructive play in order to assess conflict management skills. Some constructive and non-constructive outcomes may also be defined in such scenarios, where constructive outcomes include sharing, turn taking, trading, playing together, and other consensual settlement (e.g., rock-paper-scissors) and nonconstructive outcomes include win, loss and stalemate scenarios~\cite{shen2018stop}.

Emotion regulation plays a crucial role in conflict resolution and is defined as a process by which individuals influence which emotions they have, when they have them, and how they experience and express them~\cite{costa2018regulating}. Approaches that target emotion regulation for conflict resolution use models similar to those of affect. These may utilize physiological signals (average heart rate, mean inter-beat interval, and heart rate variability) as well as questionnaires to extract self-reported emotion states and evaluations of an interaction~\cite{costa2018regulating}. Another useful measure is group affective balance, which is defined as the balance of positive and negative affect within a group interaction~\cite{jung2016coupling}. Once again, this also requires the elicitation of self-reported affect ratings from the interactants themselves. Other approaches for emotion regulation have also been investigated. For example, a study found that by making participants listen to their own voice feedback in real time but in calmer tones (low pitch), participants' self-perception of their voice could influence their emotion regulation abilities~\cite{costa2018regulating}.

\revised{Jung et. al.~\cite{jung2015using} showed that when conflict arises in a group due to personal violations, a robot functioning as an emotional regulator could assist in managing the conflict. It did so by drawing attention to the violation, identifying it as inappropriate, and then adding a normative statement to stay positive in a humorous manner to alleviate tension.}

\revised{Approaches to personalize robot-mediated social interactions may benefit from an insight into conflict management processes within groups. Given that interventions that are issued immediately after a negative trigger are most likely to be effective~\cite{andersson1999tit}, such an insight could enable the robot to take timely action to diffuse tension in the group and up-regulate interpersonal dynamics.}

\section{Discussion and conclusion}
\label{sec:discussion-conclusion}
Human understanding models discussed in the previous section \revised{can sometimes lean towards} capturing individual human states. This may be seen in the use of physiological signals or visual indicators like facial expressions to capture affect in individual interactants. Audio cues, like the number of interruptions in speech or changes in tone, seem to better capture interpersonal aspects of social interactions. 
We discuss the implications of using individual-level states to evaluate group-level understanding in this section.

\revised{Existing HHRI research has investigated several important aspects of group dynamics, as detailed in the previous section. Based on our review of these models, we have identified some directions of future work that require further investigation. Firstly, the modeling approaches described in the previous section often focus on analyzing an interaction at the individual level, such that individual interactant states are first determined, and which may then be combined to form a collective measure. In such studies, the group may be viewed as a combination of its components.
However, group-level analysis views a group as a unified system with properties that cannot be fully understood by piecemeal examination~\cite{forsyth2018group}. This implies that when individuals merge into a group, something new is created and that the new product itself has to be the object of study, making the whole greater than the sum of the parts. Such a perspective requires further exploration in the field of HHRI and could leverage the findings from other fields, such as social psychology, where the study of groups as units has been conducted extensively~\cite{kenny1985separating, goldstone2009collective, eby1997collectivistic}.}

Additionally, the individual-level analysis raises questions about how a collective metric can be obtained from individual ones. As an example, consider an interaction between three children where two children unite to bully the third child. In such a case, we may visually observe positive affect from the facial expressions of two out of three interactants. \revised{A simple measure may favor the majority and lead us to believe that this interaction is a largely positive one, when, in fact, it is not.} Such scenarios call for a more nuanced analysis of social interactions that incorporate multimodal group-level features as well.

Secondly, existing research utilizing facial expression-based affect classification can often assume that the outwardly expressions of emotions reflect internal human states, such that a smile on a face is interpreted as an internal state of happiness. This approach does not take into account factors such as social expectations that prevent people from expressing their true feeling states in social situations~\cite{safdar2009variations}. We highlight this point, not to call for an accurate assessment of internal human states but to emphasize the need to analyze human interactions from an interpersonal rather than intrapersonal perspective. For example, two friends sharing their experiences of grief may outwardly display 
\revised{facial expressions or postures that may be interpreted as negative}, but this "negative" experience may, in fact, serve a positive social function by strengthening the bond between the two friends and bringing them closer together~\cite{clark1996some}. 

Therefore, rather than being viewed as indicators of internal state, emotional expressions must be viewed as actions that can shape people’s relational orientation towards each other \cite{van2010interpersonal}. For example, expressions of anger or hostility can push people away, whereas expressions of joy or sadness can bring people together~\cite{jung2017affective}. These findings reiterate the need to add a third dimension of interpersonal orientation to the conventional valence-arousal model of affect, and support the proposal made by Jung \cite{jung2017affective} to represent interpersonal behaviors along an axis ranging from affiliative behaviors (behaviors that turn people towards each other) to distancing behaviors (behaviors that turn people away from each other). 

\reconsider{This gives rise to the concept of relational affect, which is a dyadic construct that represents affective states that an individual experiences from interactions with others. According to Slaby~\cite{slaby2019relational}, relational affect ``does not refer to individual feeling states but to affective interactions in relational scenes, either between two or more interactants or between an agent and aspects of [her or his] environment''.
Relational affect is a consequence of the interaction itself in the form of an interplay of gaze, gesture, tone of voice etc.~\cite{slaby2019relational}, and it focuses on the observable expressions of affect between the interactants rather than the individual internal experiences of emotion.} An approach centered around the evaluation of relational affect within a group views the group as a whole and can, therefore, yield group-level metrics that can be directly utilized to evaluate existing group dynamics.

\revised{Lastly, personalization methods may also benefit from a group-level view of human-human interactions where a combination of individual-level and group-level evaluations may be used to provide mediation that is specifically tailored to the needs and preferences of the interactants within the context of existing group dynamics. Given the demonstrated importance of personalization in creating impactful robot-mediated interactions, as well as the role of group dynamics in determining, among other factors, team performance and satisfaction, we believe adding a group-level dimension to personalization approaches may be a promising direction for future work. Personalization techniques that incorporate both group dynamics and individual interactant preferences, may be able to adapt a number of robot behaviors. Examples of such behaviors include the frequency of prompts to the interactants, the empathy shown by the robot while asking the interactants questions, or even the intensity and valence of the facial expressions and gestures it may use while mediating.}

In conclusion, group-level models of human understanding that incorporate 
relational affect may be better positioned for human-human-robot interaction research, where these can be utilized \reconsider{by a robot to act in accordance with the interpersonal dynamics that exist within the group.} In alignment with the relational context of a social interaction, the mediation behaviors produced by a robot may be more likely to achieve a positive influence on existing group dynamics.
Such models could also offer a group-level, relational view of human-human interactions that may benefit approaches for personalization in HHRI research, 
enabling a robot to adapt to the needs of individual interactants in context of the social dynamics existing within the group.
By working towards building such models, we may move closer to achieving socially intelligent robot mediators that can understand the needs of the humans in their environments and produce behaviors that are effective 
in building stronger relationships 
between them.
\vspace{-0.08cm}
\section{Acknowledgments}
\label{sec:acknowledgements}
The authors would like to thank Malte Jung for their invaluable insights that contributed to the discussions presented in this work.

%\newpage
\bibliographystyle{ACM-Reference-Format}
\bibliography{sample-base}

\end{document}